\documentclass[letterpaper, 10 pt, conference]{ieeeconf}  

\IEEEoverridecommandlockouts                              

\usepackage{graphicx}
\usepackage{multirow}
\usepackage{algorithm}
\usepackage{subcaption}
\usepackage{soul}
\usepackage{comment}
\usepackage[noend]{algpseudocode}
\usepackage{afterpage}
\usepackage{array}
\usepackage{amsmath}
\usepackage{amsmath} 
\usepackage{balance}

\newcolumntype{C}{>{\centering \arraybackslash}m{0.1\textwidth}}

\title{\LARGE \bf
HetSwarm: Cooperative Navigation of Heterogeneous Swarm in Dynamic and Dense Environments through Impedance-based Guidance
}

\author{Malaika Zafar\textsuperscript{1}, Roohan Ahmed Khan\textsuperscript{1}, Aleksey Fedoseev\textsuperscript{1}, Kumar Katyayan Jaiswal\textsuperscript{2}, \\ P. B. Sujit\textsuperscript{2} and Dzmitry Tsetserukou\textsuperscript{1}%
\thanks{\textsuperscript{1}The authors are with the Intelligent Space Robotics Laboratory, Center for Digital Engineering, Skolkovo Institute of Science and Technology, Moscow, Russia. 
\tt \{malaika.zafar, roohan.khan, aleksey.fedoseev, d.tsetserukou\}@skoltech.ru}
\thanks{\textsuperscript{2}The authors are with the Multi-Robot Autonomy Laboratory,  Department of Data Science and Engineering, IISER Bhopal, Bhopal, India. 
\tt \{kumar20, sujit\}@iiserb.ac.in}
}
\begin{document}

\maketitle
\thispagestyle{empty}
\pagestyle{empty}

\begin{abstract}
With the growing demand for efficient logistics and warehouse management, unmanned aerial vehicles (UAVs) are emerging as a valuable complement to automated guided vehicles (AGVs). UAVs enhance efficiency by navigating dense environments and operating at varying altitudes. However, their limited flight time, battery life, and payload capacity necessitate a supporting ground station. To address these challenges, we propose HetSwarm, a heterogeneous multi-robot system that combines a UAV and a mobile ground robot for collaborative navigation in cluttered and dynamic conditions. Our approach employs an artificial potential field (APF)-based path planner for the UAV, allowing it to dynamically adjust its trajectory in real time. The ground robot follows this path while maintaining connectivity through impedance links, ensuring stable coordination. Additionally, the ground robot establishes temporal impedance links with low-height ground obstacles to avoid local collisions, as these obstacles do not interfere with the UAV's flight. 

Experimental validation of HetSwarm in diverse environmental conditions demonstrated a 90\% success rate across 30 test cases. The ground robot exhibited an average deviation of 45 cm near obstacles, confirming effective collision avoidance. Compared to the Conflict-Based Search (CBS) algorithm, our approach enables agents to navigate within 25 cm of obstacles, whereas CBS maintains a minimum clearance of 73 cm, highlighting our method’s efficiency in utilizing space in real-time. Extensive simulations in the Gym PyBullet environment further validated the robustness of our system for real-world applications, demonstrating its potential for dynamic, real-time task execution in cluttered environments.


\end{abstract}


\begin{figure}[htbp]
\centering
\vspace{0.2cm}
\includegraphics[width=0.95\linewidth]{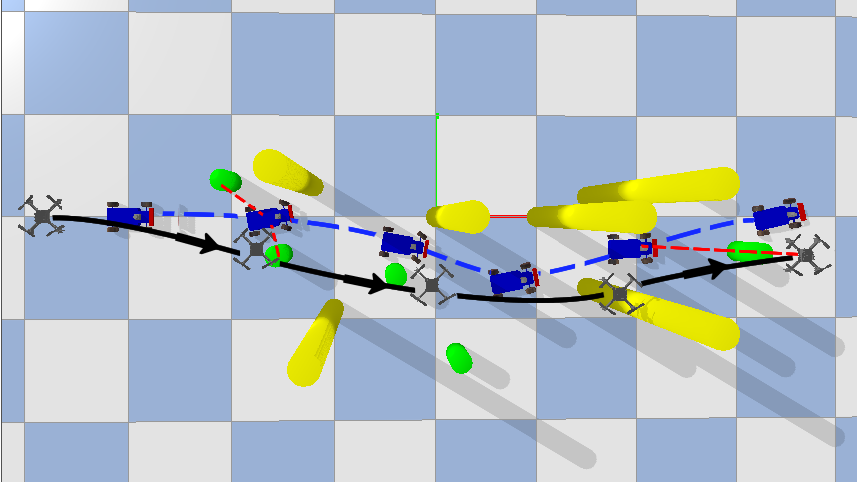} 
\caption{HetSwarm generates paths in a dense environment where the UAV (black line) navigates towards the goal and the mobile robot (blue dashed line) follows the UAV by maintaining an impedance link connection (red dashed line).}
\vspace{-0.4cm}
\label{exp_setup_1}
\vspace{-0.2cm}

\end{figure}

\section{Introduction}
With the development of automation and artificial intelligence, the logistics and warehouse management sectors have experienced a significant shift toward more efficient and automated solutions capable of decreasing the detection errors during inventory \cite{Lopez_2023}. The tasks of indoor and last-mile delivery commonly relied on AGVs and mobile robots, for example, systems proposed by Paolanti et al. \cite{Paolanti_2017} and Motroni et al. \cite{Motroni_2024}, due to their high payload capacity and localization precision. 

However, UAVs have received wide attention in logistics due to their advantages of fast speed and efficient vertical navigation. For example, an inventory management system incorporating a swarm of mini-drones proposed by Cristiani et al. \cite{Cristiani_2020} with a generic architecture for UAV-based inventory management. The segmentation model for precise position estimation of objects in inventory was proposed by Yoon et al. \cite{Yoon_2023}. 
More recently, heterogeneous systems were proposed to leverage the benefits of UAV mobility and the precise positioning of ground robots. For example, a team of mobile and aerial robots was proposed by Kalinov et al. \cite{Kalinov_2020} for real-time barcode detection and scanning using Convolutional Neural Networks (CNN). The designed approach improved the UAV's localization using scanned barcodes and ground stations as landmarks in a real warehouse with low-light conditions. 



While heterogeneous systems may improve both the precision and time of logistics, their navigation suffers from the additional density of the swarm morphology. 
This paper introduces a HetSwarm setup that combines a drone and a ground robot in a leader-follower configuration, with the drone acting as the leader while the ground robot follows them through impedance link \cite{Hogan_1984} connectivity. 

In HetSwarm, the drone uses an APF path planner to navigate to target positions while avoiding obstacles in a highly dense environment. Moreover, the ground robot maintains its connection with the drone using impedance linkages. Additionally, the ground robot makes additional impedance links with low-height obstacles that are out of the range of the drone in order to work effectively in an obstacle-dense environment. The custom PID Path Follower was made for the ground robot so that it can follow the intended path with fewer deviations. This collaborative system is also capable of functioning in dynamic environments. 

Furthermore, the heterogeneous setup significantly enhances efficiency for logistics tasks. A drone
has the ability to carry only a limited payload due to its size and battery constraints, whereas the ground robot can carry heavier payloads, thereby improving overall performance. Additionally, if the drone’s battery is depleted, it can land on the ground robot for recharging, ensuring uninterrupted operation.

\section{Related Works}

Logistics and warehouse management have become crucial components of the supply chain, with businesses demanding more efficient delivery solutions. In recent years, swarms of heterogeneous and homogeneous agents have been employed for this purpose. 

Controlling a large swarm of UAVs imposes a complex task. Several recent works propose to introduce humans in the loop for complex decision-making, e.g., Abdi et al. \cite{Abdi_2023}  with EMG-based gesture control and Khen et al. \cite{Khen_2023} proposing gesture recognition and machine learning. However, most of the systems rely on autonomous drones, with paths of vehicles being planned through various routing approaches. Batinovic et al. \cite{APF_LiDAR} utilized the APF method to address the path planning challenge for aerial robots operating in an unknown environment, focusing on ensuring safe trajectory execution and avoiding intricate obstacles using a LiDAR sensor. In a similar vein, Yu et al. \cite{Yu_2023} proposed an innovative distributed control algorithm that integrates the APF technique within a virtual leader formation scheme, coupled with a switching communication network.
Malopolski et al. \cite{GroundRobot} presented an autonomous mobile robot for transport tasks in warehouses; a drive mechanism was proposed for surface and rail navigation and an elevator for vertical movement. However, the lack of aerial capabilities limits its ability to access hard-to-reach areas and navigate dense environments, reducing adaptability in multi-level spaces. Cooperative UAV and AGV teams previously were extensively explored for rescue missions; for example, Salas et al. \cite{Salas_2021} proposed a strategy based on the exploration of an unknown environment through the collaboration of the UAV equipped with a monocular camera and the differential-drive AGV. 
Zhura et al. \cite{Zhura_2023} investigated the impact of UAVs in heterogeneous mapping and the navigation of a quadruped robot. Sales et al. \cite{Sales_2023} developed a highly scalable and low-cost multirobot system for inventory management composed of pairs with a micro-UAV and a ground mobile robot. Hajkarim et al. \cite{Hajkarim_2024} developed a new mission planning optimization method based on reinforcement learning and multi-agent rollout policy optimization for coverage missions involving uncrewed aerial systems and ground vehicles to minimize the mission planning time and heterogeneous route length. Castro et al. \cite{UAV-UGVs} proposed the strategy to assist the cooperation of a heterogeneous robot team that involves two AGVs and one UAV. The robots operate in a partially known dynamic environment where they exchange information among themselves and perform their task of aerial and ground inspections. The decentralized control barrier function was proposed by Bhatia et al. \cite{Bhatia_2024} as a solution for distributed global connectivity maintenance for a multi-agent system.

Tsykunov et al. \cite{Tsykunov_2019} first suggested the concept of SwarmTouch that utilized virtual impedance links introduced by Hogan \cite{Hogan_1984} for control of the homogeneous swarm of drones. Fedoseev et al. \cite{Fedoseev_2022} further explored this concept by analyzing different impedance link topologies in the swarm formation. Khan et al. \cite{SwarmPath} proposed a leader-follower approach using an APF path planner and impedance controller for a multi-drone homogeneous system, allowing agents to plan the path and navigate to the target in an unknown yet static environment. However, the above-mentioned approaches could not work in a dynamic environment. Additionally, the ability to operate over extended periods due to limited battery life was not previously addressed.
To overcome the individual limitations of aerial and ground robots, heterogeneous swarm systems have been used. Darush et al. \cite{SwarmGear} implemented links in a heterogeneous swarm as a formation control method between a leader octocopter and a follower swarm of micro-drones, potentially able to dock to the leader drone \cite{Karaf_2023}. Chen et al. \cite{Chen_2025} developed a heterogeneous robot system for target search and navigation, consisting of a UAV and an AGV for rescue missions in unknown environments. 

The HetSwarm system, inspired by the previous work of SwarmPath and SwarmGear, introduces a new agile and safe path planner for heterogeneous systems in dynamic and cluttered environments, relying on guiding UAVs and heterogeneous impedance links to preserve formation control. 

\section{Heterogeneous Swarm Technology}

\subsection{System Overview}

\begin{figure*}
      \centering
      \includegraphics[width=0.8\textwidth]{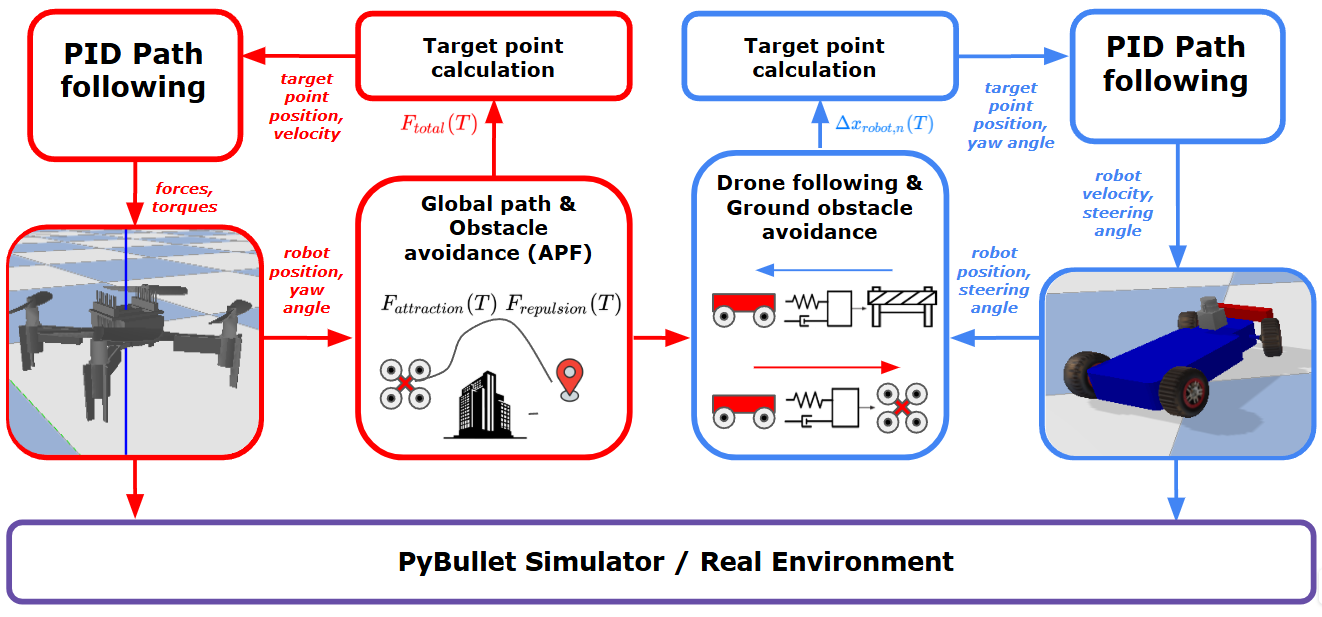}
      \caption{System architecture and the pipeline of HetSwarm for global and local collision avoidance.}
      \label{system arch}
   \end{figure*}

The HetSwarm system in Fig. \ref{system arch} consists of a drone and a ground robot working together in a dynamic environment. The drone generates and continuously updates its path using an APF planner, while the ground robot follows the drone’s path via impedance links. Additionally, these links also help the ground robot to navigate around smaller obstacles since these obstacles are not in the range of a drone. The drone handles navigation while globally avoiding obstacles, while the robot ensures obstacle-free movement while providing a landing platform to the drone in case of reloading or recharging. States of both the agents are controlled by their custom PID controller for accurate path following. This approach enables efficient collaboration in real-time, especially in densely packed dynamic environments. 

\subsection{Artificial Potential Fields for Global Path Generation}
In order for the leader drone to navigate efficiently around the obstacles while setting the path toward the goal, we applied the APF planning algorithm \cite{APF}. The algorithm generates a virtual force that attracts the drone towards the target and a repulsive force that pushes it away from obstacles. An optimized trajectory is generated using the combination of these two forces. The equations for the APF planner are as follows \cite{SwarmPath}:
\begin{equation}
F_{\text{total}} = F_{\text{attraction}} + F_{\text{repulsion}} \label{eq:total_force},
\end{equation}
where
\begin{align*}
F_{\text{attraction}}(d_{\text{g}}) &= k_{\text{att}} \cdot d_{\text{g}}, \\
F_{\text{repulsion}}(d_{\text{o}}) &= 
\begin{cases} 
0 & \text{if } d_{\text{o}} > d_{\text{safe}} \\
k_{\text{rep}} \cdot \left( \frac{1}{d_{\text{o}}} - \frac{1}{d_{\text{safe}}} \right) & \text{if } d_{\text{o}} \leq d_{\text{safe}},
\end{cases}
\end{align*}
where $d_{\text{g}}$ and $d_{\text{o}}$ are the distances from the drone to the goal and to the obstacle, respectively, $k_{\text{att}}$ and $k_{\text{rep}}$ are the attraction and repulsion coefficients, respectively.

\subsection{Impedance Controller}
Once the path of the drone was planned, to ensure a smooth connecting mechanism, its connection with the ground robot was established using an impedance controller. These impedance links help the ground robot to stay connected with the leader drone and follow its path toward the target location. 

\subsubsection{Connectivity of Mobile Robot with Drone}
In this configuration, the follower ground robot position is coupled with the leader drone APF trajectory, which serves as the guiding trajectory through a mass-spring-damper system and creates a virtual impedance link between the drone and the ground robot. These impedance links provide a smooth connection among the agents. The links are established using a second-order differential equation of mass-spring-damper, which is given as in \cite{SwarmGear}:

\begin{equation}
m\Delta\ddot{x} + d\Delta\dot{x} + k\Delta x = F_{\text{ext}}(t),
\label{eq:dynamic_equation}
\end{equation}
where $\Delta x$ is the difference between the current and desired mobile robot position and $F_{\text{ext}}(t)$ is the virtual external force applied as an input from the leader drone, $m$ is the virtual mass of a link, $d$ is the damping coefficient of the virtual damper, and $k$ is the virtual spring constant. 

\subsubsection{Connectivity of Mobile Robot with Ground Obstacles}
The APF trajectory generated by the leader drone does not account for smaller obstacles. To avoid collisions between the mobile robot and obstacles, we enabled the ground robot to establish additional links with the low-height obstacles while simultaneously disconnecting from the drone, ensuring a collision-free path:

\begin{equation}
\Delta x_{robot,n} = k_{impF} \cdot r_{imp},
\label{eq:deflection_eqn}
\end{equation}
where $r_{imp}$ is the radius of the local deflection region around the obstacle, $k_{impF}$ is the force coefficient adjusted to the ground robot's average velocity, and $n$ is the number of mobile robots. This equation ensures a collision-free trajectory by applying a repulsive displacement proportional to the obstacle’s influence radius $r_{imp}$, redirecting the robot away from potential collisions. The coefficient \(k_{\text{impF}}\) adjusts the strength of this deflection based on the robot's velocity, ensuring timely and stable avoidance.

\subsection{Dynamic Environment}
HetSwarm introduces a novel approach to path planning and coordination using a heterogeneous robot team with a UAV and a mobile robot. The system adjusts dynamically by continuously updating the drone's APF path based on real-time environmental changes. As obstacles shift, the drone adjusts its trajectory, and the setup allows the mobile robot to adapt to the changes. Additionally, the mobile robot can also dynamically react to obstacles in its path, ensuring it can navigate through unpredictable or cluttered spaces. This adaptability allows HetSwarm to operate efficiently in dynamic environments, ensuring both agents can respond to new challenges as they arise.

\section{Experimental Evaluation}

\subsection{Experimental Setup in Gym Pybullet Environment}
\begin{figure}[htbp]
\centering
\includegraphics[width=0.95\linewidth]{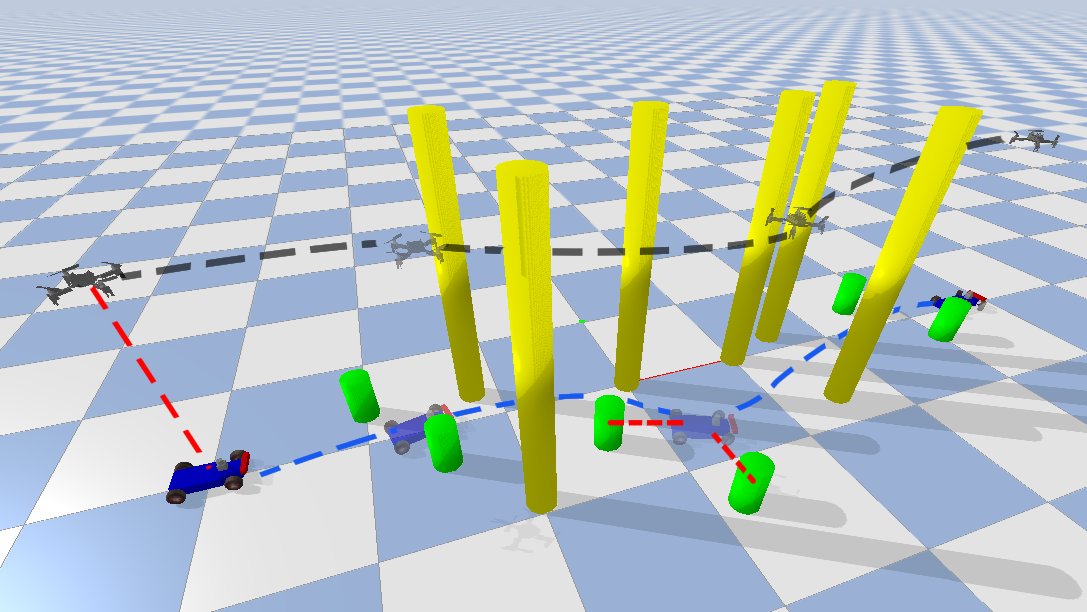} 
\caption{Experimental setup in the Gym Pybullet environment showing the trajectory of the drone and mobile robot under a dense environment.}
\label{exp_gym}
\vspace{-0.2cm}

\end{figure}

A custom simulation environment has been developed in the Gym PyBullet environment \ref{exp_gym}, incorporating the dynamics of both a drone and a mobile robot. The environment allows the visualization of multiple obstacles, and the obstacles can be moved dynamically. A custom PID controller has been implemented for the mobile robot that helps it to follow the path by adjusting its velocity and steering angles, while the drone's PID controller ensures stable and controlled drone flight. This setup creates realistic interactions between the drone and mobile robot in dynamic environments, making it suitable for testing heterogeneous robots in real-world-like conditions. The drone utilizes the APF planner to generate and adjust its path in real-time; simultaneously, the mobile robot maintains the connectivity with the drone and ground obstacles using impedance links. 
To evaluate the adaptability and robustness of the system, experiments were performed on multiple cases. In a dynamic environment, three cases were performed from sparse to dense scenarios; cases were run thirty times to check the algorithm's robustness and repeatability of performance. Similarly, in static environments, experiments were run under varying conditions to do a detailed analysis.

\subsection{Results}
\subsubsection{Results in static environment}
\begin{figure*}
\centering
\includegraphics[width=0.68\textwidth]{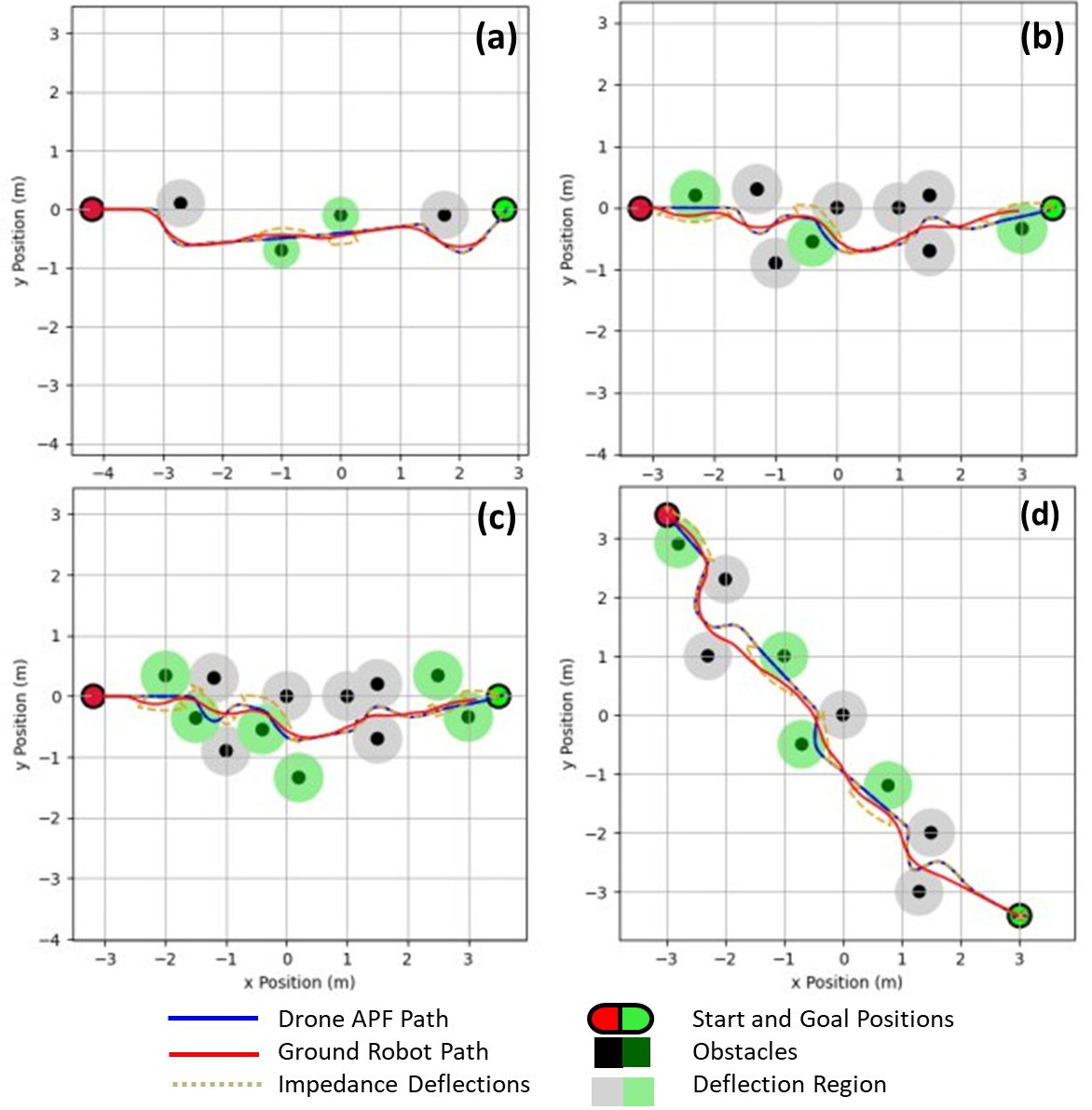}
\caption{Simulation results of two different cases: sparse (a, b) to dense (c, d). Each case is formed with two different initial and target positions. Black dots are the obstacles for both drones and mobile robots, green dots are the ground obstacles deflected only by the mobile robot. Gray and light green-colored circles show the safe deflection regions for both agents.}
\label{all_cases}
\end{figure*}

As the leader, the drone generates its path using the APF planner, while the mobile robot follows this path by maintaining impedance links, resulting in a distinct trajectory shown in Fig. \ref{all_cases}. 

Furthermore, the mobile robot avoids obstacles encountered along its path, leading to impedance deflections that are also illustrated in the figure. The green circles represent ground-level obstacles, and the graph demonstrates that the drone's trajectory remains straight, passing over these obstacles while the mobile robot effectively navigates around them. Fig.\ref{all_cases}a-c represent three cases with the same starting and goal positions of the agent, demonstrating how the algorithm effectively handles both sparse and dense scenarios. Furthermore, by changing the initial and target positions, different path planning scenarios can be tested, as shown in Fig.\ref{all_cases}d.

\subsubsection{Results in dynamic environment}
\begin{figure*}
\centering
\includegraphics[width=0.68\textwidth]{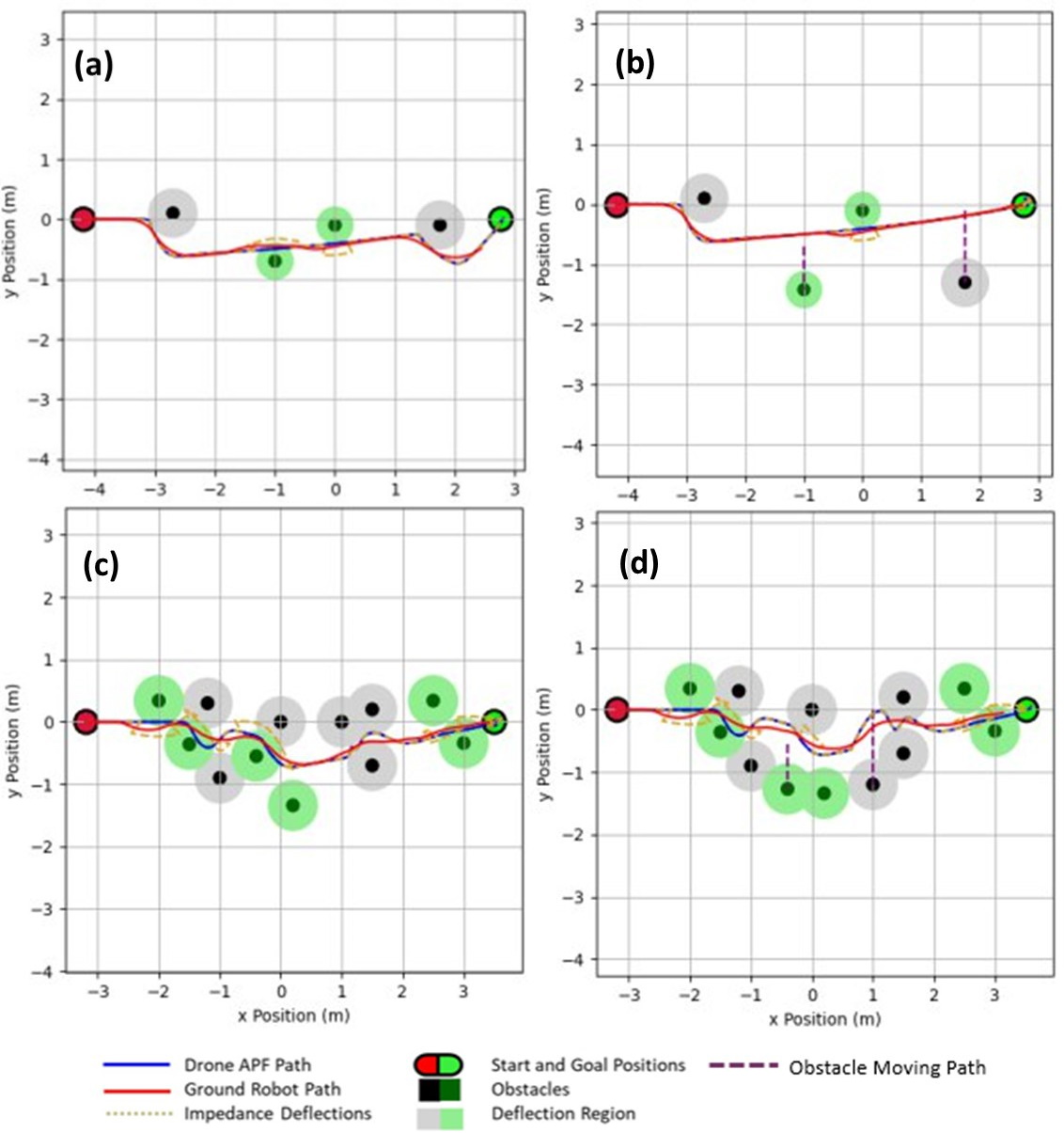}
\caption{Simulation examples of dynamic obstacle avoidance in sparse (a, b) and dense (c, d) environments. Dashed purple lines are the obstacle trajectories. Black dots are the obstacles for both robots, and green dots are the ground obstacles deflected only by the mobile robot. Gray and light green-colored circles show the safe deflection regions for both agents.}
\label{moving obs}
\end{figure*}

Real-time simulation was also performed on dynamic environments involving the continuous movement of one ground obstacle and one obstacle that was deflected by both the drone and mobile robot. Fig. \ref{moving obs} shows the tests that were performed in two main cases with sparse (Fig. \ref{moving obs}a) and dense (Fig. \ref{moving obs}b) environments. The change of path can be observed in the region where the movement of the obstacle takes place. The same behavior is observed in a cluttered environment, shown in Fig.\ref{moving obs}c and Fig.\ref{moving obs}d. The results revealed the adaptable behavior of a heterogeneous swarm under multiple environmental conditions. 

\subsubsection{Error between path planning and ground truth in simulation environment}
Since the planning is conducted in a real-time environment, the drone's position exhibits negligible errors. However, deviations are observed in the mobile robot's impedance path due to its additional responsibility of avoiding ground-level obstacles. These deviations are particularly noticeable in regions where ground obstacles are present, as the mobile robot must adjust its trajectory to navigate around them. This behavior is consistent with the system's design, where the mobile robot prioritizes obstacle avoidance while maintaining connectivity with the drone. Table \ref{Trajectory Deviation} shows the deviations of actual mobile robot positions from the planned positions.


\begin{table}[h!]
    \centering
    \caption{Error between Path Planning and Actual Performance in Simulation Environment}
    \renewcommand{\arraystretch}{1.2} 
    \begin{tabular}{|c|c|c|c|c|} 
        \hline
        \textbf{CASE NO.:} & a & b & c & d \\
        \hline
        \textbf{Ground Robot Deviations (m)} & 0.45 & 0.45 & 0.45 & 0.43 \\
        \hline
    \end{tabular}
    \label{Trajectory Deviation}
\end{table}

Fig. \ref{all_cases} shows that mobile robot deviations are higher around the ground obstacles; however, they are still maintained in the safe operating region, which shows the reliability of the path planning system and the robot's ability to handle environmental challenges, despite the presence of densely packed obstacles.

\subsubsection{Velocity Distribution along the trajectory}
\begin{figure}[htbp]
\centering
\includegraphics[width=1\linewidth]{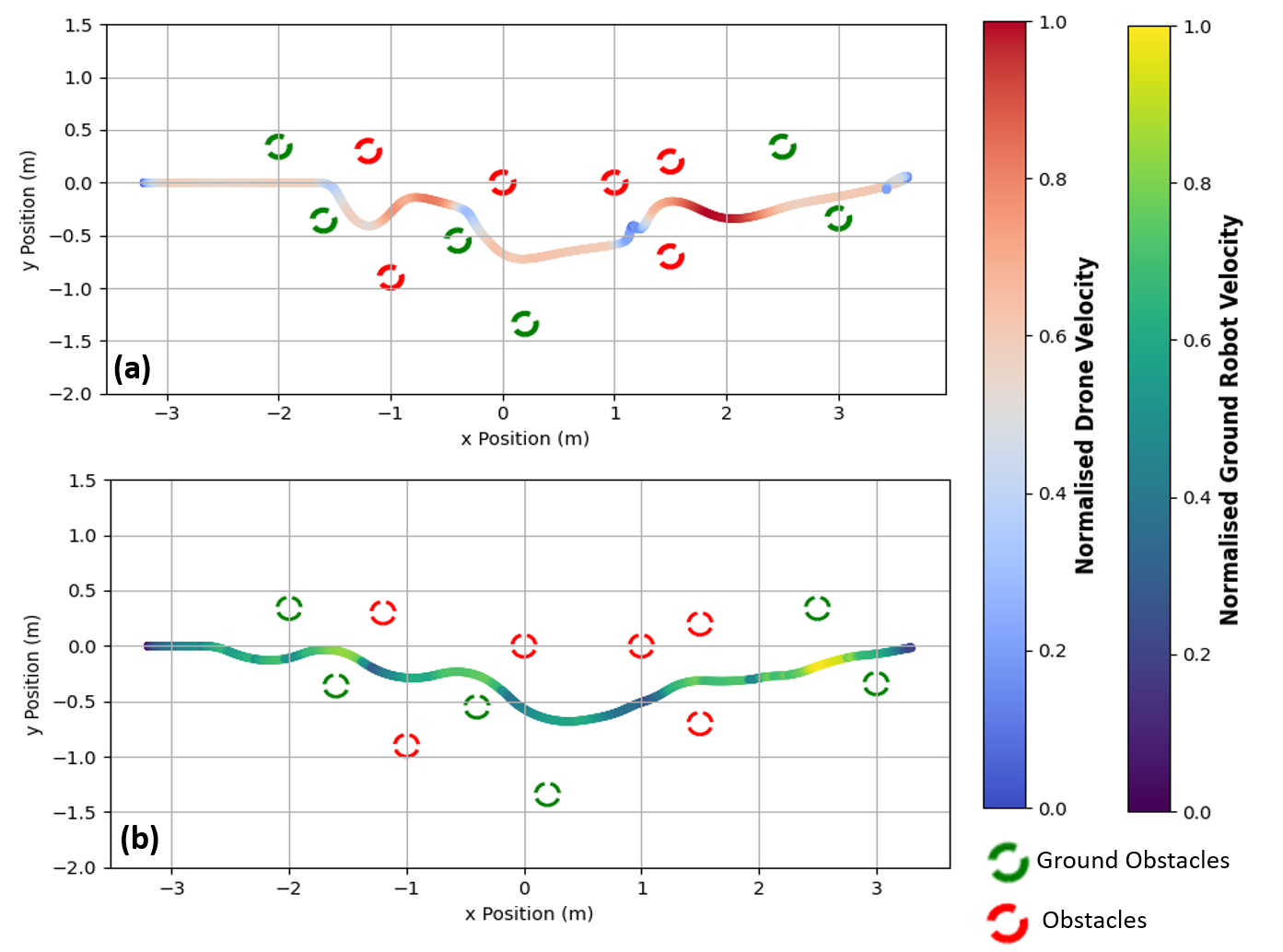} 
\caption{Visualization of Velocity Variations along the Agents Trajectory}
\label{velocity}
\vspace{-0.2cm}

\end{figure}

Fig. \ref{velocity} shows the distribution of velocities along the trajectory path for drone Fig. \ref{velocity}a, and ground robot Fig. \ref{velocity}b. The plot shows how significantly the velocities of each of the agents change over time, providing insights into their acceleration and deceleration corresponding to changing applied forces. Simulation shows that agents move faster with greater velocity near the region of obstacles. This behavior could be indicative of a strategy where the agent adjusts its velocity due to applied repulsive and impedance forces within the obstacle's close proximity to avoid collisions while maintaining a shortest and time-optimal trajectory.

\subsubsection{Time efficiency in task completion for multiple agents}
To check the robustness and system adaptability under varying conditions, we conducted experiments in two different environments, shown in Fig. \ref{all_cases}c, and Fig. \ref{all_cases}d, with varying starting and goal positions, both in a static and dynamic environment. Table \ref{trajectory} shows how the trajectory length and mission time change under varying environmental conditions.

\begin{table}[h!]
    \centering
    \caption{Swarm Behavior in Varying Conditions}
    \renewcommand{\arraystretch}{1.3}
    \begin{tabular}{| p{0.5cm}|c|c|c|c|} 
        \hline
        \multicolumn{5}{|c|}{\textbf {Time Efficiency along with length of trajectories of Leader-Drone}} \\ 
        \hline
        \textbf {Case} & \multicolumn{2}{c|}{\textbf{Static Environment}} & \multicolumn{2}{c|}{\textbf{Dynamic Environment}} \\
        \hline
        & \begin{tabular}[c]{@{}c@{}}Trajectory\\ length (m)\end{tabular} & \begin{tabular}[c]{@{}c@{}}Completion\\ time (s)\end{tabular} & \begin{tabular}[c]{@{}c@{}}Trajectory\\ length (m)\end{tabular} & \begin{tabular}[c]{@{}c@{}}Completion\\ time (s)\end{tabular}  \\
        \hline
        a & 8.18\ & 25.05\ & 8.02 \ & 25.03\\
        \hline
        b & 9.11\ & 25.42\ & 9.25\ & 25.87\\
        \hline
    \end{tabular}
    \label{trajectory}
\end{table}

The trajectory length along with completion time may increase or decrease depending on whether the agents encounter the obstacles. These results demonstrate the system's ability to handle real-time tasks and quickly and effectively adapt to environmental changes, making it suitable for real-world applications in logistics and warehouse management.

\subsubsection{Success Rate}
Simulations on the heterogeneous swarm were conducted under multiple environmental conditions. A total of 30 experiments were performed, out of which 27 were successful. Failures occurred when either of the two agents collided with an obstacle in the dynamic environment, particularly when the obstacle's speed exceeded that of the agents, causing it to suddenly block the agent's planned path. Therefore, the success rate, calculated as:
\begin{equation}
\text{Success Rate} = \left( \frac{\text{Successful Experiments}}{\text{Total Experiments}} \right) \times 100,
\end{equation}
was computed to be 90\%. 
The success rate shows that the system is relatively robust, with a strong ability to navigate and adapt to different dynamic conditions. 


\subsection{Comparison with Conflict-Based Search (CBS) Algorithm}
The CBS path planning algorithm can operate in both 2D and 3D environments. The drone trajectory generated by APF was compared with the 3D setup of the CBS algorithm, while the mobile robot trajectory was compared with the 2D setup of the CBS algorithm (Table \ref{CBS}).

\begin{table}[h!]
    \centering
    \caption{Trajectory Length and Minimum Obstacle Distance Comparison with Classic CBS Algorithm}
    \renewcommand{\arraystretch}{1.35}
    \begin{tabular}{|c|c|c|c|c|} 
        \hline
        \multicolumn{5}{|c|}{\textbf{Trajectory Length}} \\ 
        \hline
        \textbf{Case} & \multicolumn{2}{c|}{\textbf{2D Environment}} & \multicolumn{2}{c|}{\textbf{3D Environment}} \\
        \hline
        & \textbf{CBS} & \textbf{mobile robot} & \textbf{CBS} & \textbf{Drone} \\
        \hline
        a & 10.39\ & 9.02\ & 7.5\ & 7.86 \\
        \hline
        b & 10.39\ & 12.36\ & 7.5
        \ & 8.18 \\
        \hline
        \multicolumn{5}{|c|}{\textbf{Minimum Obstacle Distance}} \\ 
        \hline
        \textbf{Case} & \multicolumn{2}{c|}{\textbf{2D Environment}} & \multicolumn{2}{c|}{\textbf{3D Environment}} \\
        \hline
        a & 0.73\ & 0.25\ & 0.75\ & 0.38 \\
        \hline
        b & 0.72\ & 0.25\ & 0.75\ & 0.37 \\
        \hline
    \end{tabular}
    \label{CBS}
\end{table}

With similar environmental configurations, the CBS trajectory in Case 1, where there are fewer obstacles, shows that agents tend to move around obstacles, resulting in a longer path. In contrast, APF agents navigate within obstacles, leading to more efficient paths. However, in Case 2, where the environment is more densely packed with obstacles, CBS agents ignore the inner regions and move around obstacles to complete the trajectory. On the other hand, APF agents move between obstacles, encountering multiple deflections, which increase their path length. Despite this, APF excels in minimum obstacle distance; agents maintain a very close distance to obstacles while still avoiding collisions. In comparison, CBS agents maintain a minimum distance of 72 cm, reflecting a more cautious but less flexible path.


\subsection{Conclusion and Future Work}
The research presents a heterogeneous swarm system comprising a drone and a mobile robot and designed to collaborate in dynamic environments for logistics and warehouse management tasks. This research works on a leader-follower approach where a drone, being a leader, uses an APF path planner to navigate to goal positions while avoiding obstacles. Moreover, the mobile robot follows the drone using impedance links that connect the drone with the mobile robot. Moreover, the mobile robot develops additional impedance links with the ground obstacles to ensure smooth navigation toward the goal. 

By combining the two agents, the system proves to have the complementary strengths of both a drone and a mobile robot. Multiple experiments were performed on this system, achieving a success rate of 90\%. Additionally, a deviation of 45 cm was observed in the mobile robot's path near the ground obstacles region, indicating that the robot is effectively avoiding them. The current approach was later compared with the Conflict-Based Search (CBS) algorithm, which reveals that our approach enables agents to safely yet optimally navigate within 25 cm of obstacles, whereas CBS maintains a minimum distance of 73 cm. 
 
In the future, we plan to expand this system to include multiple drones and mobile robots working together in more complex environments. We also aim to test the system in real-world settings, ensuring its feasibility in practical logistics and warehouse operations. 

\section*{Acknowledgements} 
Research reported in this publication was financially supported by the RSF-DST grant No. 24-41-02039.

\bibliographystyle{IEEEbib}

\balance
\bibliography{HetSwarm}

\begin{thebibliography}{10}

\bibitem{Lopez_2023}
S.~López-Soriano and Rafael Pous,
\newblock ``Inventory robots: Performance evaluation of an rfid-based navigation strategy,''
\newblock {\em IEEE Sensors Journal}, vol. 23, no. 14, pp. 16210--16218, 2023.

\bibitem{Paolanti_2017}
Marina Paolanti, Mirco Sturari, Adriano Mancini, Primo Zingaretti, and Emanuele Frontoni,
\newblock ``Mobile robot for retail surveying and inventory using visual and textual analysis of monocular pictures based on deep learning,''
\newblock in {\em Proc. 2017 European Conference on Mobile Robots (ECMR)}, 2017, pp. 1--6.

\bibitem{Motroni_2024}
Andrea Motroni, Salvatore D’Avella, Alice Buffi, Paolo Tripicchio, Matteo Unetti, Glauco Cecchi, and Paolo Nepa,
\newblock ``Advanced rfid-robot with rotating antennas for smart inventory in high-density shelving systems,''
\newblock {\em IEEE Journal of Radio Frequency Identification}, vol. 8, pp. 559--570, 2024.

\bibitem{Cristiani_2020}
Davide Cristiani, Filippo Bottonelli, Angelo Trotta, and Marco Di~Felice,
\newblock ``Inventory management through mini-drones: Architecture and proof-of-concept implementation,''
\newblock in {\em Proc. 2020 IEEE 21st International Symposium on "A World of Wireless, Mobile and Multimedia Networks" (WoWMoM)}, 2020, pp. 317--322.

\bibitem{Yoon_2023}
Bohan Yoon, Hyeonha Kim, Geonsik Youn, and Jongtae Rhee,
\newblock ``3d position estimation of objects for inventory management automation using drones,''
\newblock {\em Applied Sciences}, vol. 13, no. 19, 2023.

\bibitem{Kalinov_2020}
Ivan Kalinov, Alexander Petrovsky, Valeriy Ilin, Egor Pristanskiy, Mikhail Kurenkov, Vladimir Ramzhaev, Ildar Idrisov, and Dzmitry Tsetserukou,
\newblock ``Warevision: Cnn barcode detection-based uav trajectory optimization for autonomous warehouse stocktaking,''
\newblock {\em IEEE Robotics and Automation Letters}, vol. 5, no. 4, pp. 6647--6653, 2020.

\bibitem{Hogan_1984}
Neville Hogan,
\newblock ``Impedance control: An approach to manipulation,''
\newblock in {\em Proc. 1984 American Control Conference}, 1984, pp. 304--313.

\bibitem{Abdi_2023}
Sydrak~S. Abdi and Derek~A. Paley,
\newblock ``Safe operations of an aerial swarm via a cobot human swarm interface,''
\newblock in {\em Proc. 2023 IEEE International Conference on Robotics and Automation (ICRA)}, 2023, pp. 1701--1707.

\bibitem{Khen_2023}
Golan Khen, Detim Zhao, and Jos\'{e} Baca,
\newblock ``Intuitive human-swarm interaction with gesture recognition and machine learning,''
\newblock in {\em Proc. 24th International Symposium on Theory, Algorithmic Foundations, and Protocol Design for Mobile Networks and Mobile Computing}, New York, NY, USA, 2023, MobiHoc '23, p. 453–456, Association for Computing Machinery.

\bibitem{APF_LiDAR}
Ana Batinovic, Jurica Goricanec, Lovro Markovic, and Stjepan Bogdan,
\newblock ``Path planning with potential field-based obstacle avoidance in a 3d environment by an unmanned aerial vehicle,''
\newblock in {\em Proc. 2022 International Conference on Unmanned Aircraft Systems (ICUAS)}, 2022, pp. 394--401.

\bibitem{Yu_2023}
Yajing Yu, Chen Chen, Jian Guo, Mohammed Chadli, and Zhengrong Xiang,
\newblock ``Adaptive formation control for unmanned aerial vehicles with collision avoidance and switching communication network,''
\newblock {\em IEEE Transactions on Fuzzy Systems}, vol. 32, no. 3, pp. 1435--1445, 2024.

\bibitem{GroundRobot}
Waldemar Małopolski and Sebastian Skoczypiec,
\newblock ``The concept of an autonomous mobile robot for automating transport tasks in high-bay warehouses,''
\newblock {\em Advances in Science and Technology Research Journal}, vol. 18, pp. 1--10, 04 2024.

\bibitem{Salas_2021}
William~Leith Salas, Luis~M. Valentín-Coronado, Israel Becerra, and Alfonso Ramírez-Pedraza,
\newblock ``Collaborative object search using heterogeneous mobile robots,''
\newblock in {\em Proc. 2021 IEEE International Autumn Meeting on Power, Electronics and Computing (ROPEC)}, 2021, vol.~5, pp. 1--6.

\bibitem{Zhura_2023}
Iana Zhura, Denis Davletshin, Nipun Dhananjaya~Weerakkodi Mudalige, Aleksey Fedoseev, Robinroy Peter, and Dzmitry Tsetserukou,
\newblock ``Neuroswarm: Multi-agent neural 3d scene reconstruction and segmentation with uav for optimal navigation of quadruped robot,''
\newblock in {\em Proc. 2023 IEEE International Conference on Systems, Man, and Cybernetics (SMC)}, 2023, pp. 2525--2530.

\bibitem{Sales_2023}
Augusto Sales, Pedro Mira, Ana Maria~Nascimento, Alexandre Brandão, Martin Saska, and Tiago Nascimento,
\newblock ``Heterogeneous multi-robot systems approach for warehouse inventory management,''
\newblock in {\em Proc. 2023 International Conference on Unmanned Aircraft Systems (ICUAS)}, 2023, pp. 389--394.

\bibitem{Hajkarim_2024}
Sina M.~H. Hajkarim, P.~B. Sujit, and Prathyush~P. Menon,
\newblock ``Optimized mission planning for heterogeneous uncrewed vehicle teams,''
\newblock {\em IEEE Access}, vol. 12, pp. 147387--147399, 2024.

\bibitem{UAV-UGVs}
Gabriel Castro, Tatiana Santos, Fabio Andrade, José Lima, Diego Haddad, Leonardo Honório, and Milena Faria~Pinto,
\newblock ``Heterogeneous multi-robot collaboration for coverage path planning in partially known dynamic environments,''
\newblock {\em Machines}, vol. 12, pp. 200, 03 2024.

\bibitem{Bhatia_2024}
Pranjal Bhatia, Sayan~Basu Roy, P.B Sujit, Luis~Mejias Alvarez, and Aaron McFadyen,
\newblock ``Decentralized connectivity maintenance for multi-agent systems using control barrier functions,''
\newblock in {\em Proc. 2024 International Conference on Unmanned Aircraft Systems (ICUAS)}, 2024, pp. 955--962.

\bibitem{Tsykunov_2019}
Evgeny Tsykunov, Ruslan Agishev, Roman Ibrahimov, Luiza Labazanova, Akerke Tleugazy, and Dzmitry Tsetserukou,
\newblock ``Swarmtouch: Guiding a swarm of micro-quadrotors with impedance control using a wearable tactile interface,''
\newblock {\em EEE Trans. Haptics}, vol. 12, no. 3, pp. 363–374, July 2019.

\bibitem{Fedoseev_2022}
Aleksey Fedoseev, Ahmed Baza, Ayush Gupta, Ekaterina Dorzhieva, Riya~Neelesh Gujarathi, and Dzmitry Tsetserukou,
\newblock ``Dandeliontouch: High fidelity haptic rendering of soft objects in vr by a swarm of drones,''
\newblock in {\em Proc. 2022 IEEE International Conference on Systems, Man, and Cybernetics (SMC)}, 2022, pp. 1078--1083.

\bibitem{SwarmPath}
Roohan~Ahmed Khan, Malaika Zafar, Amber Batool, Aleksey Fedoseev, and Dzmitry Tsetserukou,
\newblock ``Swarmpath: Drone swarm navigation through cluttered environments leveraging artificial potential field and impedance control,'' 2024.

\bibitem{SwarmGear}
Zhanibek Darush, Mikhail Martynov, Aleksey Fedoseev, Aleksei Shcherbak, and Dzmitry Tsetserukou,
\newblock ``Swarmgear: Heterogeneous swarm of drones with reconfigurable leader drone and virtual impedance links for multi-robot inspection,'' 2023.

\bibitem{Karaf_2023}
Sausar Karaf, Aleksey Fedoseev, Mikhail Martynov, Zhanibek Darush, Aleksei Shcherbak, and Dzmitry Tsetserukou,
\newblock ``Morpholander: Reinforcement learning based landing of a group of drones on the adaptive morphogenetic uav,''
\newblock in {\em Proc. 2023 IEEE International Conference on Systems, Man, and Cybernetics (SMC)}, 2023, pp. 2507--2512.

\bibitem{Chen_2025}
Yun Chen and Jiaping Xiao,
\newblock ``Target search and navigation in heterogeneous robot systems with deep reinforcement learning,''
\newblock {\em Machine Intelligence Research}, vol. 22, no. 1, pp. 79–90, Jan. 2025.

\bibitem{APF}
Haoyang Li,
\newblock ``Robotic path planning strategy based on improved artificial potential field,''
\newblock in {\em Proc. 2020 International Conference on Artificial Intelligence and Computer Engineering (ICAICE)}, 2020, pp. 67--71.

\end{thebibliography}

\end{document}